\documentclass{ecai} 

\usepackage{latexsym}
\usepackage{amssymb}
\usepackage{amsmath}
\usepackage{amsthm}
\usepackage{booktabs}
\usepackage{enumitem}
\usepackage{graphicx}
\usepackage{color}

\newcommand{\BibTeX}{B\kern-.05em{\sc i\kern-.025em b}\kern-.08em\TeX}

\begin{document}


\begin{frontmatter}


\paperid{2331} 


\title{Forecasting Events in Soccer Matches Through Language}


\author[A,B]{\fnms{Tiago}~\snm{Mendes-Neves}\thanks{Corresponding Author. Email: tiago.neves@fe.up.pt}} 
\author[B]{\fnms{Luís}~\snm{Meireles}} 
\author[A,C]{\fnms{João}~\snm{Mendes-Moreira}} 

\address[A]{Faculdade de Engenharia Universidade do Porto}
\address[B]{Nordensa Football}
\address[C]{LIAAD - INESC TEC}


\begin{abstract}
This paper introduces an approach to predicting the next event in a soccer match, a challenge bearing remarkable similarities to the problem faced by Large Language Models (LLMs). Unlike other methods that severely limit event dynamics in soccer, often abstracting from many variables or relying on a mix of sequential models, our research proposes a novel technique inspired by the methodologies used in LLMs. These models predict a complete chain of variables that compose an event, significantly simplifying the construction of Large Event Models (LEMs) for soccer.
Utilizing deep learning on the publicly available WyScout dataset, the proposed approach notably surpasses the performance of previous LEM proposals in critical areas, such as the prediction accuracy of the next event type. This paper highlights the utility of LEMs in various applications, including match prediction and analytics. Moreover, we show that LEMs provide a simulation backbone for users to build many analytics pipelines, an approach opposite to the current specialized single-purpose models.
LEMs represent a pivotal advancement in soccer analytics, establishing a foundational framework for multifaceted analytics pipelines through a singular machine-learning model.
\end{abstract}

\end{frontmatter}


\section{Introduction}
Despite its global popularity and economic impact, soccer has lagged behind other sports in leveraging data analytics for insights. This gap stems from the inherent dynamism of the game, exceeding other team sports due to the larger number of players and constant interactions. As a result, soccer presents a vast, fertile ground for advancements in sports analytics, mainly through leveraging the power of artificial intelligence.

Recent developments in AI, specifically the emergence of foundational models capable of learning the full spectrum of a game and facilitating insightful interpretation, offer exciting possibilities for revolutionizing soccer analytics. Large Events Models (LEMs) \cite{mendes_neves_towards_LEM} exemplify this paradigm.
LEMs represent a novel approach in the field of sports analytics, particularly tailored for understanding and predicting sequences of events in soccer. Unlike traditional models that rely on isolated statistical metrics, LEMs draw inspiration from the architecture of Large Language Models (LLMs), which have demonstrated remarkable success in natural language processing. By conceptualizing a soccer match as a series of events — each akin to a "word" in a sentence — LEMs utilize deep learning techniques to predict the next event in the sequence based on the context provided by previous events. This methodology allows for a dynamic and contextually rich analysis of sports games, capturing events' fluid and unpredictable nature as they unfold during a match.

The core innovation of LEMs lies in their ability to predict discrete outcomes and generate probabilistic forecasts for every aspect of an event, enabling the generation of entire sequences of game events and offering a comprehensive tool for game simulation and analysis. This ability is achieved through sequential deep learning models, each responsible for forecasting a specific aspect of the event. By feeding data into a series of predictive models, LEMs can simulate various possible future scenarios of a match from any given moment, thereby providing insights into likely game developments. 
Still, LEMs have several limitations.

\begin{figure}[h]
  \centering
  \includegraphics[width=\linewidth]{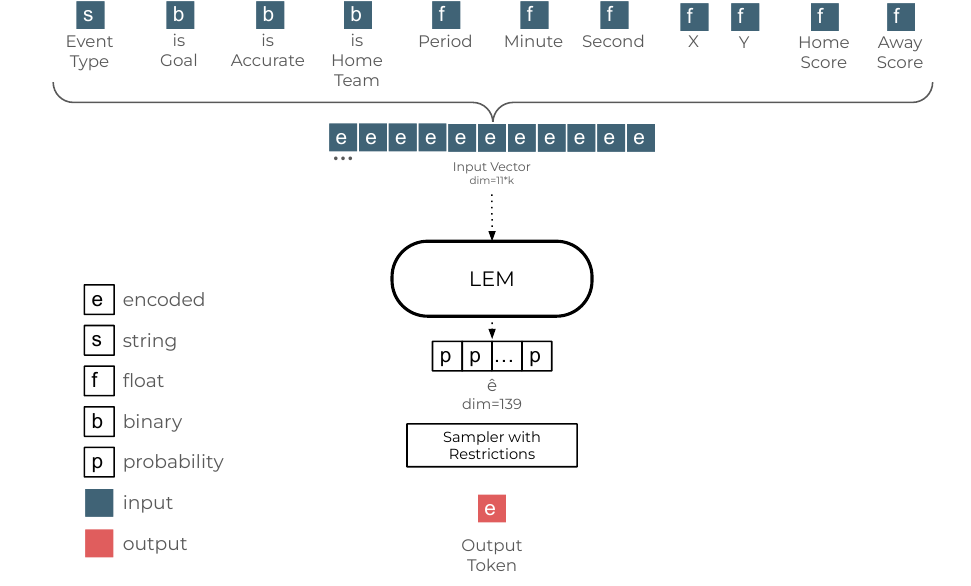}
  \caption{A schematic representation of our proposal. In blue, we have the set of inputs used to build the input vector, passed through the LEM model to infer the probabilities of each token. To make a prediction, the probabilities go through a sampler with restrictions to avoid hallucinations, i.e., predicting unrealistic values.}
  \label{fig:llm_overview}
\end{figure}

We present a novel approach to building LEMs that addresses the limitations of existing LEMs, depicted in Figure \ref{fig:llm_overview}. Instead of relying on sequential inferences from multiple models, we draw inspiration from the core methodologies of language models. By tokenizing soccer event data, we enable a single model to learn the "language" of soccer events effectively. This single model performs sequential inferences, with each inference corresponding to a token representing a part of an event. Our model captures the most relevant aspects of soccer events with seven inferences, achieving superior performance in forecasting key variables compared to conventional LEMs.

Furthermore, we explore the extensive potential of LEMs. By leveraging their ability to model complex event sequences, LEMs can unlock valuable insights applicable to various domains within the soccer analytics ecosystem. These include:

\begin{itemize}
    \item 
\textbf{Match Prediction:} LEMs can predict the likelihood of future events, empowering users to make informed decisions based on a comprehensive understanding of game dynamics.
    \item 
\textbf{Sports Analytics Tools:} Integrating LEMs within advanced analytics tools can enable automated and in-depth analysis of match strategies, player performance, and tactical trends.
    \item 
\textbf{Simulation and Scenario Planning:} LEMs can generate realistic simulations of hypothetical scenarios, aiding coaches and analysts in evaluating potential strategies and planning future matches.
\end{itemize}

By demonstrating the versatility and efficacy of our LEMs across diverse use cases and through comparison with prominent frameworks, we aim to solidify their position as a valuable tool for enriching the landscape of soccer analytics and generating a deeper understanding of the sport.

This paper is organized as follows:
\begin{itemize}
    \item Section \ref{sec:related_work} provides an overview of the current state of LEMs, including their limitations and areas to improve.
    \item Section \ref{sec:language-based_approach} details our approach to build a single-model LEM, including datasets used, encoding, formulation, and hyperparameters.
    \item Section \ref{sec:experiments} shows several experiments detailing experimental results and applications.
    \item Section \ref{sec:conclusion} presents the concluding remarks.
\end{itemize}

\section{Related Work} \label{sec:related_work}

Soccer has witnessed a growing trend in data-driven analysis, aiming to uncover patterns and predict future outcomes. From injury prediction \cite{kakavas_artificial_2020} to transfer values \cite{mchale_estimating_2023}, there are several applications where machine learning can help soccer teams and players improve their performance.

However, these applications are driven through specialized models. For each new application, a new model needs to be built from scratch, which is time-consuming and prone to errors. To solve this, simulation environments \cite{kurach2020google} and approaches \cite{mendes_neves_datadriven_simulator_2021} have been proposed. However, they are still limited in scope.

LEMs \cite{mendes_neves_towards_LEM} draw inspiration from Large Language Models (LLMs), which are trained on vast datasets and can be fine-tuned for specific tasks. By treating sequences of game events as "sentences" of actions, where the model predicts the next "word" describing the event based on the "context" of previous events, LEMs provide a generative framework to make evaluations in soccer. As soccer matches are sequential, this framework allows the simulation of soccer matches from start to end. LEMs employ a sequential forecasting method to align and predict multiple variables more accurately, predicting variables in a staged approach as follows:
\begin{itemize}
    \item 
\textbf{Model Type:} This stage predicts the type of the next event.
    \item 
\textbf{Model Accuracy:} This stage assesses the accuracy of the upcoming event and predicts whether it will result in a goal.
    \item 
\textbf{Model Data:} This final stage predicts the location of the next event, the time until this event occurs, and whether the home team will be involved.
\end{itemize}

In each stage, the LEM generates a probability distribution for each variable. For example, the output from the Model Type provides probabilities for each potential event type. Similar to the approach used in LLMs, our model samples from these probabilities to determine the next event type. Therefore, each forecast is probabilistic, implying that the output might vary for the same contextual input based on the predicted probabilities.

LEMs show promise regarding progress towards a general approach \cite{mendes_neves_towards_LEM}. It shows significant progress over previous approaches \cite{simpson_seq2event_2022}, which were limited in scope, mostly due to only using a small portion of offensive events. By (1) extending the covered events and (2) significantly accelerating the inference process, the original LEM proposal enabled large-scale simulations of soccer matches. These simulations can be used for many use cases, including those stated before injury prediction, valuing player contribution, and forecasting probabilities.

However, the current iteration of LEMs still faces significant challenges that limit their practical applicability. While the models advanced over previous iterations by enabling complete event forecasts, crucial issues remain. A major shortcoming lies in their reliance on three separate models performing three inferences, encompassing seven variables. This complex architecture leads to several drawbacks:

\begin{enumerate}
    \item 
\textbf{Complex Architecture:} LEMs use three different multi-layer perceptron models to predict different aspects of a soccer event. One model predicts the event type, then a different model predicts the accuracy of the event, and only then is the model able to infer spatiotemporal aspects of the event.
    \item 
\textbf{Synchronization Challenges:} Independent training can lead to incompatibility between the models and convergence to divergent local optima.
    \item 
\textbf{Partial Information:} Inferences for certain variables may lack the full context required for optimal accuracy.
    \item 
\textbf{Hyperparameter Tuning Complexity:} Tuning independent hyperparameters for each model significantly complicates the optimization process.
\end{enumerate}

The approach proposed in this paper provides a substantial improvement for all of these drawbacks: (1) We use a single model to forecast the whole event, which simplifies architecture, removes synchronization challenges, and reduces the number of hyperparameters requiring tuning, and (2) each aspect of the event is calculated using the full context of the previous forecasts, increasing the amount of information, and consequently leading to an increase in performance.

LLMs can predict complete sentences: they do not need a model to forecast nouns and another to forecast adjectives. The ability of LLMs to learn general knowledge and generate coherent data inspired the exploration of approaches similar to LLMs to solve the problem of forecasting the next event in a soccer match.

Applying techniques used in LLMs to other areas is not novel. LLMs have transcended text generation, demonstrating success in diverse domains like music composition \cite{oord2016wavenet} and physical simulation \cite{jain2022zeroshot}.

Generative artificial intelligence is still in the early stages of soccer. Our proposal provides a ground foundation to advance LEMs toward real applications in soccer by providing a significant improvement in terms of model accuracy and ease of implementation.

\section{A Language-based Approach to LEMs} \label{sec:language-based_approach}

\subsection{Data}
We use the Public Wyscout dataset, which contains data from Wyscout V2 API \cite{pappalardo_public_2019}. The dataset includes data for the 2017/18 season of the five most valuable leagues in European football: England, Spain, Germany, Italy, and France. This dataset does not represent the most up-to-date data standards available in the industry, but it is one of the most up-to-date publicly available datasets. We considered merging datasets from different sources. However, the standards between providers have substantial differences, which would incur more limitations to our work.
We extract the key features highlighted in Figure \ref{fig:llm_overview} for each event. 

We use the France, Germany, and Italy leagues on our training set, leaving the complete seasons of England and Spain for testing. Although not including data from the testing league in the training set leads to worse performance, we opted to leave these leagues entirely out of the dataset so they could be used for season-long application development using our models without the risk of overfitting.
This limitation is exclusive of the publicly available datasets and should

The 11 features used per event are the following:
\begin{itemize}
    \item Event Type: Categorical variable indicating the event, e.g, pass, shot, take on, tackle.
    \item isGoal: Binary variable indicating whether the event is related to a goal.
    \item isAccurate: Binary variable indicating if the action was successful.
    \item isHomeTeam: Binary variable indicating if the home team performed the action.
    \item Period: Integer variable describing the current half being played, i.e., 1st or 2nd.
    \item Minute: Integer variable with the current minute.
    \item Second: Integer variable with the current second.
    \item X: Integer variable with the X coordinate of the event.
    \item Y: Integer variable with the Y coordinate of the event
    \item Home Score: Integer variable indicating the current score of the home team.
    \item Away Score: Integer variable indicating the current score of the away team.
\end{itemize}

Some of the features could be merged since they are co-dependent: movement across the x coordinate affects movement across the y coordinate. However, combining these two variables would lead to a large increase in our output space (50+ times larger) and, consequently, a worse computational performance. Therefore, the variables are predicted sequentially, with the x variable being predicted first due to its higher importance relative to the y coordinate.

The model does not include features like team and player identification since they substantially increase the input and output space. There are over 2000 players in our dataset, increasing the model size. Including specific team or player information data on LEMs remains an area for future research.

It is important to note that event data contains several limitations. Therefore, our model will also extend some of these limitations. Examples of these limitations are (1) the absence of off-the-ball data, which represents over 98\% of the time players spend on the pitch, (2) biases regarding annotation edge cases, (3) manual annotation of continuous variables are imprecise, among others.

\subsection{Leveraging Language Models for Event Prediction}

Soccer can be abstracted to Markov chains \cite{rudd_2011,decroos_actions_2019}. A soccer match is a sequence of events performed by the players on the field. This sequential nature of soccer makes it a fertile ground for Markov-based approaches to forecasting many aspects of the game, e.g., next event \cite{simpson_seq2event_2022,mendes_neves_towards_LEM}, action value \cite{decroos_actions_2019}. It is easy to draw the parallel between generating the next events and generating the next word. As LLMs use the previous words as context to forecast the next word, LEMs use the previous events as context to forecast the next event.

Several aspects of LLMs are relevant to our approach. LLMs provide a basis for sequential generation, which is crucial - we do not want to forecast the next event, but rather a chain of future events and their outcomes. Sequential generation enables LEMs to generate full soccer matches, enabling the extraction of multiple metrics from the data. LLMs are also very good at understanding previous contexts. Context in soccer is key. The next event should reflect the game's current state, including several long-term aspects such as the current score.

In abstraction, LLMs tokenize and encode the text to facilitate the learning problem. Tokenization is breaking down a large text or data into smaller units called tokens \cite{9075398}. For example, the word "learning" can be split into the tokens "learn" and "ing". This step facilitates the algorithm's learning process in multiple aspects. The "ing" token is repeated across multiple instances in the English language - decoupling the "ing" from "learning" will enable the reutilization of the same token across all gerund words. Then, the tokens need to be encoded into numbers, which deep learning models can use.

For our proposal, we do not perform tokenization of event data. The vocabulary in event data is minimal, with only 140 words used for the selected data points. There is no semantic aspect to our data other than sequentially. Therefore, in this use case, tokenization does not significantly impact our models' performance.

The same rationale can be applied to the encoding. The simplicity of our approach makes advanced approaches, like embeddings \cite{9075398}, inefficient. Employing an ordinal encoding approach is more practical. The tokens are encoded in the following way:
\begin{itemize}
    \item numeric values 0 to 100 occupy the first 101 positions of the encoder.
    \item categorical values regarding action types occupy the following 37 encoder positions, from the most to least frequent.
    \item <PERIOD\_OVER> and <GAME\_OVER> tokens occupy the following two encoder positions.
    \item the <NaN> token occupies the last position of the encoder.
\end{itemize}

An example of the resulting data format is presented in Table \ref{tab:encoded_data}, extracted from the sample of event data presented in Table \ref{tab:event_data}.

\begin{table*}[h]
    \centering
    \begin{tabular}{ccccccccccc}
        \hline
        Event Type & IsGoal & IsAccurate & IsHome & Period & Minute & Second & X & Y & HomeScore & AwayScore \\
        \hline
        ground\_attacking\_duel & False & True & False & 2 & 47 & 11 & 93 & 14 & 0 & 2 \\
        ground\_defending\_duel & False & False & True & 2 & 47 & 11 & 7 & 86 & 0 & 2 \\
        simple\_pass & False & True & False & 2 & 47 & 12 & 97 & 36 & 0 & 2 \\
        shot & True & True & False & 2 & 47 & 14 & 87 & 43 & 0 & 3 \\
        \hline
    \end{tabular}
    \caption{This table shows a sample of the raw event data extracted from the Wyscout dataset.}
    \label{tab:event_data}
\end{table*}

\begin{table*}[h]
    \centering
    \begin{tabular}{ccccccccccc}
        \hline
        Event Type & IsGoal & IsAccurate & IsHome & Period & Minute & Second & X & Y & HomeScore & AwayScore \\
        \hline
        102 & 0 & 1 & 0 & 2 & 47 & 11 & 93 & 14 & 0 & 2 \\
        103 & 0 & 0 & 1 & 2 & 47 & 11 & 7 & 86 & 0 & 2 \\
        101 & 0 & 1 & 0 & 2 & 47 & 12 & 97 & 36 & 0 & 2 \\
        116 & 1 & 1 & 0 & 2 & 47 & 14 & 87 & 43 & 0 & 3 \\
        \hline
    \end{tabular}
    \caption{This table presents the encoded version of the event data from Table \ref{tab:event_data}, using ordinal encoding. This format is suitable for machine learning algorithms.}
    \label{tab:encoded_data}
\end{table*}

Our approach fixes several limitations of previous proposals: (1) it simulates all action types, overcoming the limitation of only being able to simulate the attacking portion of the game \cite{simpson_seq2event_2022}, (2) it generates events from a singular model, severely simplifying the architecture and preserving complete sequentiality of the generation process.

The first limitation is fixed by modeling the use of all event types in our model, enriching the amount of captured event dynamics. The second limitation is overcome by our methodology, using a similar approach to language generation models to address the problem of event forecasting.

\subsection{Formulation}

LEMs use a set ($S$) of previous events ($e$) to forecast the next event, as formalized in Equations \ref{eq:state_general} and \ref{eq:state_k1}. This set of events has size $k$, defining how many previous events are used to forecast the next event.

\begin{equation} \label{eq:state_general}
S_k = [e_{-1}, e_{-2}, …, e_{-k}]
\end{equation}

\begin{equation} \label{eq:state_k1}
S_1 = [e_{-1}]
\end{equation}

Our approach models the probability ($p$) for each possible token ($t$). Therefore, we can obtain the probability vector ($\hat{p}$) according to Equation \ref{eq:p_from_state}.

\begin{equation} \label{eq:p_from_state}
\hat{p} = f(S_k)
\end{equation}

The predicted $\hat{p}$ vector is then used to sample the next event token through a Multinomial sampler. The sampler contains restrictions regarding which part of the event is being predicted. The restrictions act as a diversity control system. If it predicts the first token within an event, corresponding to the event type, then only the probabilities related to event type tokens are passed to the sampler. The restrictions are required to ensure consistent outputs from the models. As the model will be used for millions of inferences and simulations, any residual probability inconsistent with the event data format can derail the prediction of future events, making the simulation unusable. 

However, our approach can only forecast one token at a time. To solve this problem, we need to provide the model with information about the current composition of the event. The model needs to be informed about which position of the event tuple is being predicted. The predicted elements are added to the set $S$ to achieve this. This way, the model is informed about which event positions were already predicted and can better forecast the next position within the event. For modeling purposes, we pad the flattened vector with the <NaN> token, which the model can use to determine which predictions are valid for the next token.

There are two exceptions in the prediction of the next event, which is not the same as the input:
\begin{itemize}
    \item 
Current scores for home and away teams are computed externally, avoiding the burden of extra inferences to calculate a deterministically computable variable.
    \item 
The Period, Minute, and Second variables are transformed into a TimeElapsed variable. This transformation shortens the number of inferences required by two. All three variables are then deterministically computed from the previous values of the variables, the predicted Event Type, and Time Elapsed variables.
\end{itemize}

\subsection{Deep Learning}

Recent advances in deep learning architectures enabled large LLM performance increases, such as RNNs \cite{hochreiter_long_1997} and Transformers \cite{radford_improving_nodate}. However, these architectures substantially increase the number of learnable parameters in the neural networks. In LLMs, the increase in the number of parameters can be easily compensated with an increase in computational power and larger datasets. GPT-2 models \cite{radford2019language} use 40GB of data to train a 1.5B parameter model.

Due to the constraints in publicly available soccer event data, we used a multi-layer perceptron architecture for our model. After pre-processing, the available event data in our dataset is 100 MB. Although we see significant gains to be made using more advanced architectures, the size of the data available to us is insufficient to explore these approaches thoroughly.

Our architecture was determined by trial and error. We increased the size of the model layers in powers of 2 until we found no tangible gain over the previous iteration. We repeated the process for 2 and 3 layers. The final architectures are detailed as follows:
\begin{itemize}
    \item 
Hidden Layers: The model contains three layers with 512 neurons each, with the lite model containing two layers with 256 neurons each. For the k=1 model, these architectures contain around 600k and 100k parameters, respectively. We use ReLU as the activation function. Although the architecture seems overparameterized when compared with the number of features available, it is important to note that we have a much larger output space.
    \item 
Learning Rate: Initiated at a value of 0.001, the learning rate undergoes dynamic adjustment each epoch, following a cosine annealing schedule \cite{huang2017snapshot}. This approach aids in stabilizing the learning process over epochs. 
    \item 
Epochs: The training was conducted over 50 epochs. This number of epochs was chosen to allow sufficient time for the model to converge.
    \item 
Loss Function: Binary Cross-Entropy Loss (BCELoss) was used as the criterion for training the model. This choice is particularly suitable for binary classification problems, offering a probabilistic approach to classification tasks.
    \item 
Optimizer: The Adam optimizer \cite{kingma_adam_2017} was employed for its efficiency in handling sparse gradients and adaptive learning rate capabilities.
\end{itemize}

One of the LEMs' main purposes is to enable the large-scale simulation of soccer games. For such, the user only needs to iteratively query the model, while arranging the game's current state to be updated by the new predictions. 

Moreover, the LEM is implemented so that the only limiting factor for the number of parallel simulations is the amount of available GPU memory. This architecture allows the parallel simulation of millions of soccer matches concurrently, enabling simulation-based insights at a large scale.

In some games, the current event has frequent long-term impacts on the outcome of matches. Due to soccer's low-scoring nature, a few factors can significantly impact the current game status, completely changing the context of the match. This is a great feature for our models - as soccer contains a lot of "reset points", such as set pieces, or whenever the ball goes out-of-bound, a new sequence is started. With the start of a new sequence, the cumulative error of our predictions has a soft reset. While some variables, such as the current score, have a significant impact on the following outcome, these resets allow our model to rebalance its predictions in cases where the model starts to hallucinate.

\subsection{Implementation and Reproducibility}
For the implementation, we used PyTorch running on a NVIDIA 3060 12GB. The code required to reproduce this work is available at 
\url{https://github.com/nvsclub/LargeEventsModel}.

\section{Experiments} \label{sec:experiments}
\subsection{Training the LEMs}

For this paper, we trained three models:
\begin{itemize}
    \item \textbf{K=1}: a model that only uses the previous event as the game state, similar to the original proposal of LEM.
    \item \textbf{K=1s}: a lite model trained with a significantly smaller number of parameters (around 6x fewer parameters than K=1).
    \item \textbf{K=3}: a model using the last 3 previous events as the game state.
\end{itemize}

The performance of the three models is presented in Table \ref{tab:lem_performance_metrics}. We also report the results of a baseline (BL), which uses the majority class as a prediction in classification and uses the average values of the variable in regression.

\begin{table}[ht]
\centering
\caption{Comparative performance metrics of baseline and LEM predictive models across various variables and metrics. The table showcases accuracy (ACC) and F1-scores for categorical outcomes (Event Type, Goal, Accurate, Home), as well as Mean Absolute Error (MAE) and R-squared (R2) values for continuous variables (Time Elapsed, X, Y). This comprehensive analysis highlights the improvements in predictive accuracy and precision when utilizing different configurations of the LEM model compared to the baseline (BL). The LEM column corresponds to the original proposal \cite{mendes_neves_towards_LEM}.}
\label{tab:lem_performance_metrics}
\begin{tabular}{llccccc}
\hline
\textbf{Variable} & \textbf{Metric}    & \textbf{BL} & \textbf{LEM} & \textbf{K=1} & \textbf{K=1s} & \textbf{K=3} \\ \hline
Type &                                                           
ACC           & 40.8\%            & 55.7\%       & 57.5\%       & 57.3\%            & \textbf{62.2\%}       \\
 & F1           & 0.24              & 0.5          & 0.52         & 0.52              & \textbf{0.57}         \\ \hline
Goal & 
ACC           & 99.7\%            & \textbf{99.8\%}       & \textbf{99.8\%}       & \textbf{99.8\%}            & \textbf{99.8\%}       \\
 & F1           & 0              & \textbf{0.87}         & 0.68         & 0.68              & 0.68         \\ \hline
Accurate & 
ACC           & 67.8\%            & 81.7\%      & 82.7\%       & 82.5\%            & \textbf{82.8\%}      \\
 & F1           & 0                 & 0.69         & \textbf{0.87}         & \textbf{0.87}              & \textbf{0.87}         \\ \hline
Home &
ACC           & 50.9\%            & \textbf{93.8\%}       & 92.1\%       & 91.5\%            & 93.6\%       \\
 & F1           & 0              & \textbf{0.94}         & 0.92         & 0.92              & \textbf{0.94}         \\ \hline

Time & MAE              & 3.1               & \textbf{1.6}          & \textbf{1.6}          & 1.7               & 1.7          \\
& R2         & 0                 & \textbf{0.55}         & 0.45         & 0.39              & 0.42         \\ \hline
X & MAE              & 21.2              & 8.5          & 6.7          & 7.4               & \textbf{6.5}          \\
& R2         & 0                 & 0.29         & 0.81         & 0.77              & \textbf{0.82}         \\ \hline
Y & MAE              & 26.5             & 15.6        & 12.1         & 12.8              & \textbf{11.4}         \\
& R2         & 0                 & \textbf{0.64}         & 0.54         & 0.50              & 0.59         \\ \hline
 
\end{tabular}
\end{table}

Our proposed LEMs significantly improve soccer event prediction accuracy compared to a traditional baseline model and the previous iteration of LEMs. Table \ref{tab:lem_performance_metrics} reveals substantial gains across various metrics, including event type, accuracy prediction, and spatial coordinates. Only in two variables, \textit{isHome} and \textit{TimeElapsed}, our proposal did not improve over the previous iteration. Moreover, some improvements are substantial: (1) event type accuracy increased by 6.5\% for the K=3 model, and (2) spatial coordinates X and Y decreased error by 24\% and 28\%, respectively.

The results regarding the Goal/Accurate variables offer a glimpse into this approach's trade-off against the original LEM proposal. The greater granularity of predictions, which are now predicted individually rather than massively, leads to increased performance in the variable predicted later, at a performance cost of the previously predicted variable. The original LEM proposal leads to better predictive performance. However, the methodology requires multiple variables to be forecasted simultaneously. This is where our current proposal allows for a significant leap. We can now forecast all variables sequentially, increasing the information available in each event prediction and making the predicted events more cohesive than the original approach.

We also ran tests to evaluate the inference speed of each algorithm. Compared to K=1, K=1s offers a 21\% increase in inference speed, while K=3 slows down inference by 62\%.

\subsection{Model Inspection}

\begin{figure}[h]
  \centering
  \includegraphics[width=\linewidth, trim=0pt 0pt 280pt 0pt, clip]{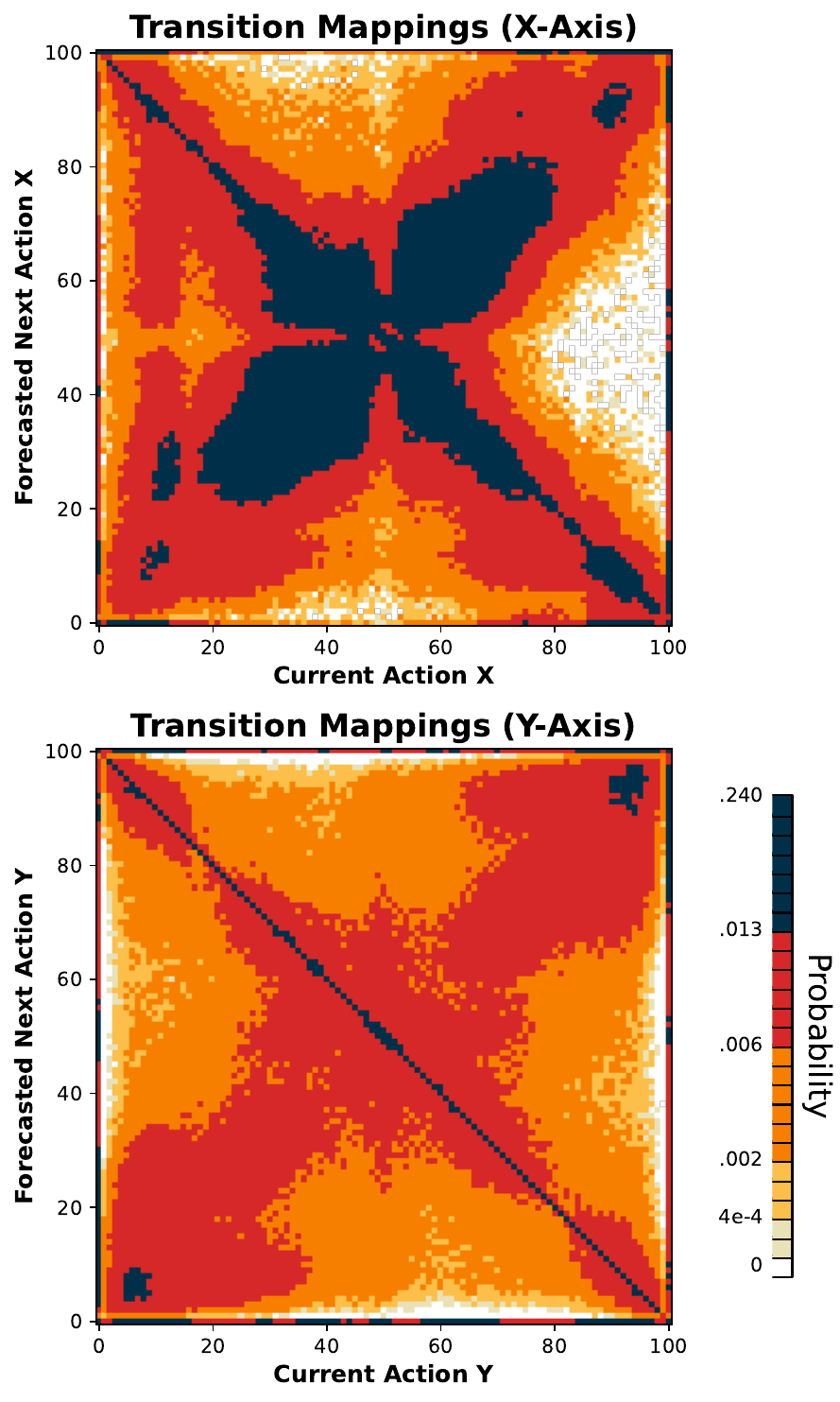}
    \caption{The probability of transitioning from current location x,y to the next location x,y. The pattern contains two behaviors: (1) the positive correlation between the current coordinates and the next coordinate, as the next event performed by the same team is expected to be close to the current event, and (2) a negative correlation caused by when the next event is performed by the opposite team, as the coordinate axis changes to the opposition's perspective.}
  \label{fig:loc_matrix}
\end{figure}

This section focuses on the model's ability to predict the next event location, a crucial aspect for understanding player movement patterns and team strategies. We evaluated the K=1 model by analyzing the learned transition matrices, representing the probability of transitioning from one location to another, presented in Figure \ref{fig:loc_matrix}.

The transition matrices exhibit significant matching, particularly from bottom left to top right. This indicates the model accurately predicts transitions.
A prominent diagonal extends from the top left to the bottom right, highlighting the high probability of possession changes. As events are always annotated from the team attacking perspective, the coordinates change to the symmetrical point when there is a change of possession between teams.

\subsection{Computing Situational Expected Goals Maps}

\begin{figure}
\begin{minipage}[b]{0.32\linewidth}
K=1s
\includegraphics[height=1.7\linewidth, trim=12cm 0pt 1.8cm 0pt, clip]{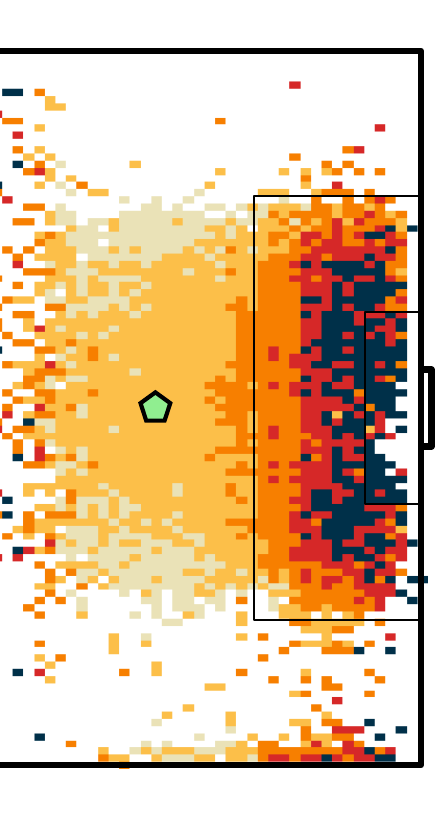}
\end{minipage}
\hfill 
\begin{minipage}[b]{0.32\linewidth}
K=1
\includegraphics[height=1.7\linewidth, trim=12cm 0pt 1.8cm 0pt, clip]{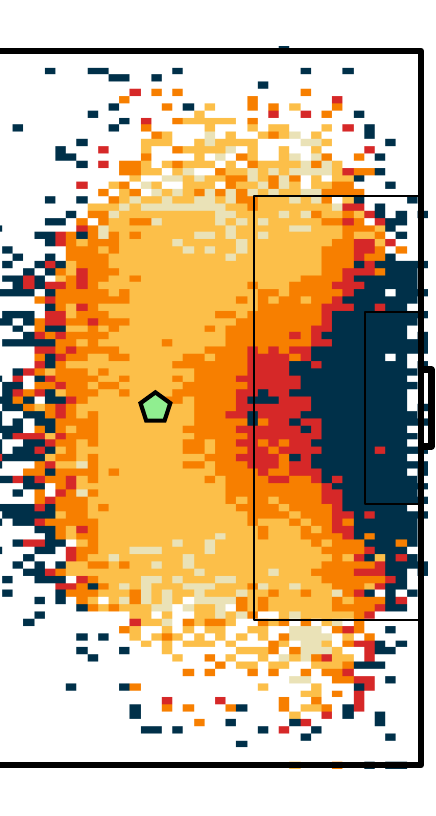}
\end{minipage}
\hfill 
\begin{minipage}[b]{0.32\linewidth}
K=3
\includegraphics[height=1.7\linewidth, trim=12cm 0pt 0pt 0pt, clip]{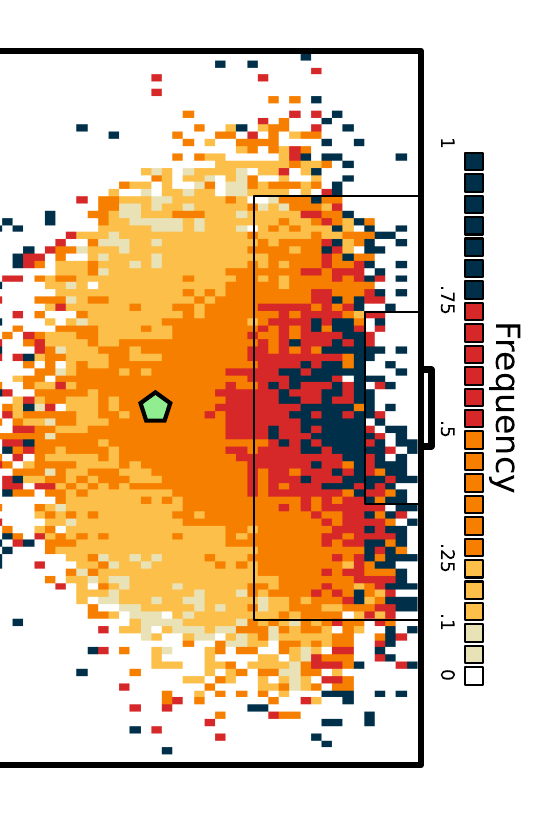}
\end{minipage}


\begin{minipage}[b]{0.32\linewidth}
\includegraphics[height=1.7\linewidth, trim=12cm 0pt 1.8cm 0pt, clip]{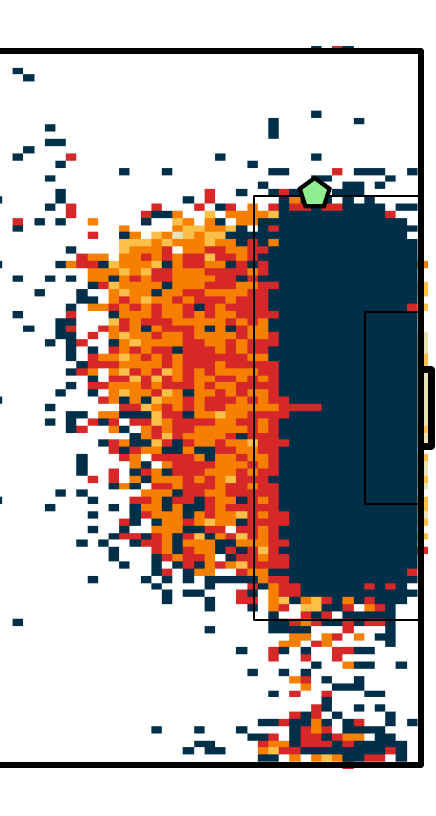}
\end{minipage}
\hfill 
\begin{minipage}[b]{0.32\linewidth}
\includegraphics[height=1.7\linewidth, trim=12cm 0pt 1.8cm 0pt, clip]{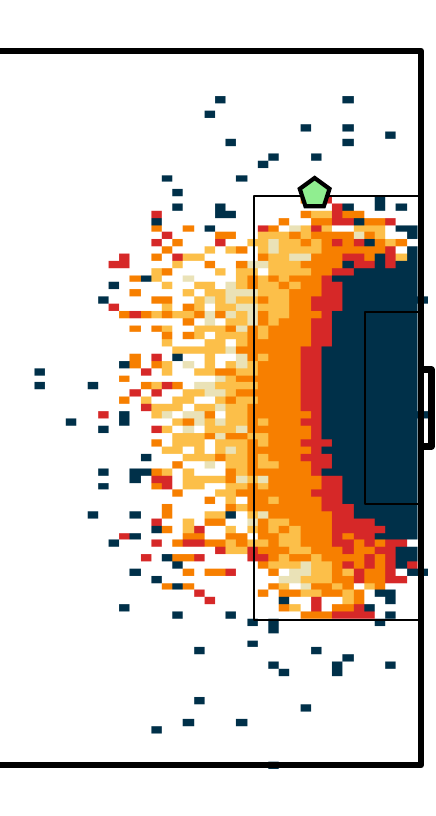}
\end{minipage}
\hfill 
\begin{minipage}[b]{0.32\linewidth}
\includegraphics[height=1.7\linewidth, trim=12cm 0pt 0pt 0pt, clip]{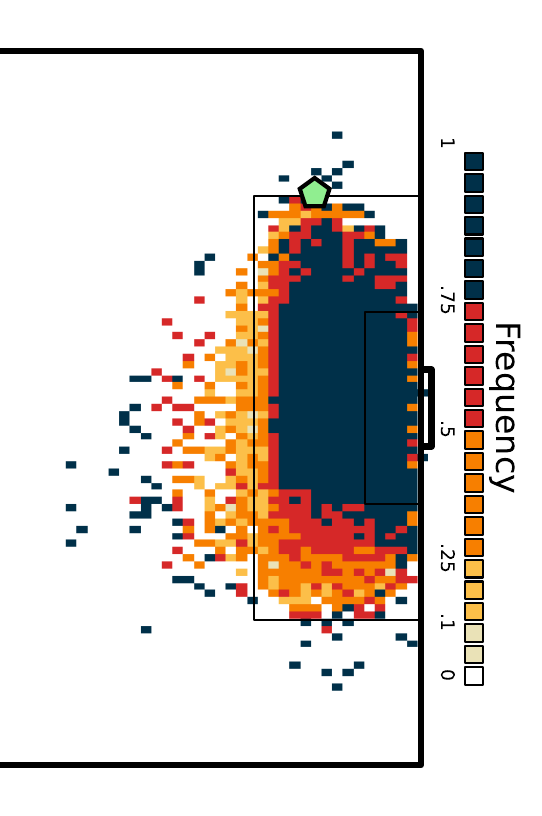}
\end{minipage}

\caption{The situational expected goals maps calculated across the different models. For each case, we simulated 1.000.000 shots for each input. Then, we calculate the percentage of shots leading to a goal for each location, which is used to plot the figures.}
\label{fig:expected_goals_situations}
\end{figure}

The expected goals metric (xG) is a widely used measure of opportunity quality in soccer. It measures the probability of a given shot situation leading to a goal. Figure \ref{fig:expected_goals_situations} shows the expected goals map from two different game states. The first game state happens after a pass in the coordinates (80, 50), while the second occurs after a cross in the coordinates (90, 80). For the K=3 model, both actions are preceded by two passes from coordinates (20, 50) and (50, 50), emulating a fast transition situation.

Several insights are visible in \ref{fig:expected_goals_situations}. First, we observe data points with a high distance from the goal and abnormal values for expected goals. These anomalies occur due to a bias in event data: when a cross goes toward the goal, it is labeled as a shot if it ends in a goal but is still labeled as a cross if it gets claimed by the keeper. This discrepancy diverges between real goal expectations and our model's goal expectations. Nonetheless, this data bias is well reported in the literature and recurring across different datasets \citep{marreiros_data-driven_2021}.

The K=1s model struggles to learn the basic patterns in data. The results are less precise than the peer models. The trade-off of lowering the number of parameters for increased inference speed affects performance substantially. 

Interesting differences arise between K=1 and K=3. As both models are large enough to learn the patterns in data, the larger context in the K=3 model learns different patterns from the K=1 model. The K=3 model has access to two previous actions that indicate progressive passes from the shooting team. This information helps the model expect a shot from a situation likely to be a fast transition. The added information increases the precision of the maps from the K=1 model that is unaware of this context.

\subsection{Forecasting Short-term Probabilities}

LEMs also can forecast probabilities at multiple time scales. In soccer, we define short-term probability as the probability of scoring a goal shortly after the event. 

\begin{figure}[h]
  \centering
  \includegraphics[width=\linewidth]{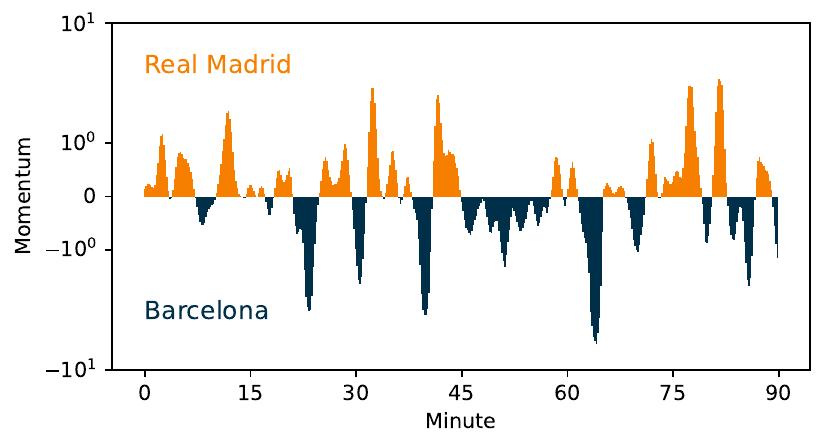}
  \caption{The visualization of a match momentum indicator built using the K=1 model. The data corresponds to the Real Madrid - Barcelona, December 23, 2017.}
  \label{fig:momentum_indicator}
\end{figure}

The most famous use case for short-term probabilities is the momentum indicators in live score apps (e.g., Sofascore). Figure \ref{fig:momentum_indicator} shows an example of a momentum indicator.

\subsection{Forecasting Long-term Probabilities}

\begin{figure}[h]
  \centering
  \includegraphics[width=\linewidth]{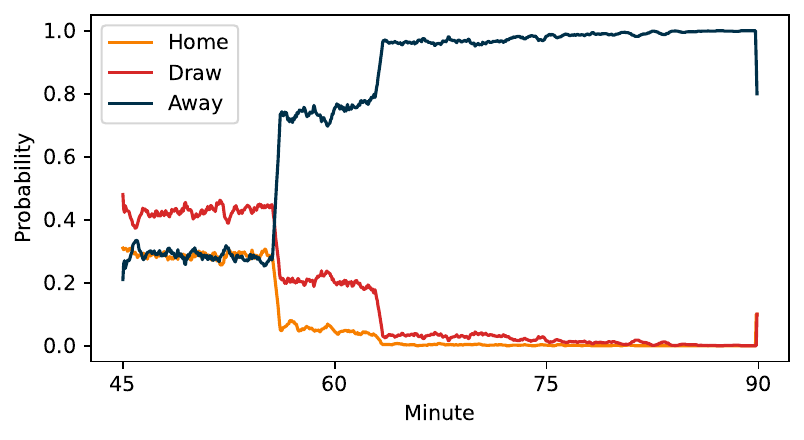}
  \caption{The in-game probabilities calculated using LEMs for the game Real Madrid - Barcelona, December 23, 2017. The second half starts with a balance in probabilities and shifts abruptly every time Barcelona scores a goal. The short-term fluctuations provoked by events in the match are also visible in the image.}
  \label{fig:overtime_probability}
\end{figure}

\begin{figure}[h]
  \centering
  \includegraphics[width=\linewidth]{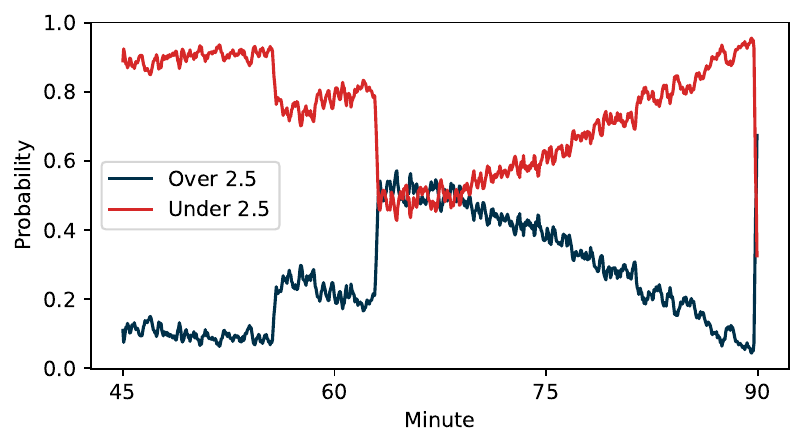}
  \caption{The in-game probabilities for whether the game will have under or over 2.5 goals, calculated using LEMs for Real Madrid - Barcelona, December 23, 2017. The probability of the current winning bet (under) increases as time passes but sharply decreases when goals are scored. After the third goal, the outcome of this market is certain, and the probabilities sharply move to 0 and 1, respectively.}
  \label{fig:overtime_under_over_probability}
\end{figure}

In contrast to short-term probabilities, long-term probabilities refer to a large time scale, usually to predict the outcome of a full match. With the simulations ran on LEMs, we can calculate the frequency of outcomes across a range of variables - everything from who wins (see Figure \ref{fig:overtime_probability}, to goals (see Figure \ref{fig:overtime_under_over_probability}, corners and others.

The forecasted probabilities for both game outcome and the probability of the game ending with over or under 2.5 goals follow the expected pattern. The current result of the match increases its probability until a score change happens. When a goal occurs, the probabilities sharply move, benefiting the team that scored.

Figures \ref{fig:overtime_probability} and \ref{fig:overtime_under_over_probability} look like they have a lot of noise. The noise represents the slight fluctuations of probabilities due to the match flow: as a team becomes likelier to score, their probabilities increase at a much smaller scale than their increase when scoring a goal. Therefore, the probability noise resembles the changes in probability between events.

One of the advantages of using LEMs for such purposes is that with the same simulations, we can calculate a wide range of probabilities. We do not need to run a new simulation for each new variable we want to forecast. We can reuse the generated event data from previous simulations and extract the probabilities.
As for the limitations, the model cannot capture the shift in probabilities due to a red card in the 63rd minute, right before Barcelona's second goal. This factor is absent from the context given to LEMs.

\subsection{VAEP}

When estimating both short-term and long-term probabilities in soccer matches, we enable calculating action values via the Valuing Actions by Estimating Probabilities (VAEP) framework \cite{decroos_actions_2019}. VAEP defines the value of an action as the resultant increase in the probability of scoring minus the increase in the probability of conceding. Using this framework, we calculated the VAEP value for the actions before the last goal in the Real Madrid - Barcelona, December 23, 2017. These results are presented in Table \ref{tab:vaep_table}. We compare VAEP scores with ST/10 values, which assess the short-term scoring impact of actions within the time frame of 10 actions, to understand their alignment. Furthermore, we examine long-term probabilities (LT/inf) and a modified version (LT*/inf) calculated with a hypothetical 0-0 score to evaluate their effectiveness in different contexts. All values use the K=1 model.

\begin{table}[h!]
\centering
\begin{tabular}{ccccccc}
\hline
\textbf{ID} & \textbf{Type} & \textbf{Team} & \textbf{VAEP} & \textbf{ST/10} & \textbf{LT/inf} & \textbf{LT*/inf} \\
\hline
912061 & pass & away & 0 & -0.011 & -0.002 & 0.002 \\
912062 & pass & away & -0.01 & -0.001 & 0.002 & -0.02 \\
912063 & pass & away & 0.01 & -0.003 & -0.002 & -0.003 \\
912064 & take on & away & 0.05 & -0.003 & 0.002 & -0.003 \\
912065 & tackle & home & - & -0.012 & 0.001 & -0.029 \\
912066 & pass & away & 0.09 & 0.109 & 0.002 & 0.18 \\
912067 & shot & away & 0.83 & 0.865 & 0 & 1.727 \\
912068 & reflexes & home & - & 0.003 & 0 & 0.001 \\
\hline
\end{tabular}
\caption{The table compares the VAEP measures of 4 different approaches. The VAEP column represents the original VAEP approach \cite{decroos_actions_2019}. ST/10 stands for short-term probability over a horizon of 10 events, which are parameters that allow a fair comparison with VAEP. LT/inf stands for long-term probability over an infinite horizon - this means that the game in simulated until the <GAME\_OVER> event type appears in the simulation. LT*/inf modifies the current score to 0-0, substantially changing the measures' context. Both LT approaches weight outcomes by the amount of points it yields to the team (3 for a win, 1 for a draw, 0 for a loss) since soccer matches do not have binary outcomes.}
\label{tab:vaep_table}
\end{table}

VAEP values closely resemble ST/10 scores for most actions, indicating their effectiveness in capturing the immediate impact on scoring opportunities.
As expected, the pass directly assisting the goal exhibits a significantly high VAEP score, reflecting its critical role in creating the scoring chance. 
Furthermore, there is an agreement on the high impact of the shot action, with both VAEP and ST/10 giving similar scores.

One action where VAEP and ST/10 disagree is the value of the dribble. The difference is explained by the better context in the original VAEP probability estimator - the estimator knows where the dribble ends, while LEMs currently do not accept this information as input. Therefore, the added information of VAEP leads to a different valuation for the dribble. 
The initial LT/inf values, considering the entire game and its outcome, underestimate the value of the final shot since the match was already decided. However, when we adjust for a hypothetical 0-0 score (LT*(inf), the value of the shot aligns more accurately with the expected VAEP impact, although using a different scale.

These results are very important from the point of view of player valuation. Currently, the most relevant metrics for player valuation are aggregating the values of their events. Therefore, by evaluating events, we are effectively assessing the players.

\section{Conclusion} \label{sec:conclusion}

This paper presented a LEM-based approach for predicting event sequences in soccer matches with remarkable results. The proposed model achieved good accuracy and surpassed previous iterations, demonstrating the effectiveness and promise of this methodology.

Future research focusing on contextual enrichment has the potential to unlock even greater accuracy and broader applicability of LEMs. Exploring advanced architectures like RNNs and Transformers can also propel this approach to new heights.
This approach is a significant step forward in developing LEMs, opening doors for groundbreaking applications in soccer research. The possibilities are vast, from enhanced tactical analysis to match prediction applications.

\begin{ack}
This work is financed by National Funds through the Portuguese funding agency, FCT - Fundação para a Ciência e a Tecnologia, within project UIDB/50014/2020.
\end{ack}


\bibliography{mybibfile}

\end{document}